\documentclass[letterpaper, 10 pt, conference]{ieeeconf}  % Comment this line out if you need a4paper

\IEEEoverridecommandlockouts                              % This command is only needed if 
\usepackage{epsfig} % for postscript graphics files
\usepackage{times} % assumes new font selection scheme installed
\usepackage{amsmath} % assumes amsmath package installed
\usepackage{amssymb}  % assumes amsmath package installed
\usepackage{gensymb}
\usepackage[utf8]{inputenc}
\usepackage{textcomp, gensymb}

\usepackage{bm}
\usepackage{url}
\usepackage{nccmath}
\usepackage{siunitx}
\usepackage{wrapfig}

\usepackage[usenames] {color}

\newcommand{\optional}[1]{}

\title{\LARGE \bf

Robotic Object Insertion with a Soft Wrist \\ through Sim-to-Real Privileged Training
}

\author{Yuni Fuchioka$^{1*}$, Cristian C. Beltran-Hernandez$^{1}$, Hai Nguyen$^{1*}$, and Masashi Hamaya$^{1}$% <-this % stops a space
\thanks{$^{1}$OMRON SINIC X Corporation, Tokyo, Japan. $^{*}$Work done as an intern.}%%
}

\begin{document}

\maketitle
\thispagestyle{empty}
\pagestyle{empty}

%%%%%%%%%%%%%%%%%%%%%%%%%%%%%%%%%%%%%%%%%%%%%%%%%%%%%%%%%%%%%%%%%%%%%%%%%%%%%%%%
\begin{abstract}
This study addresses contact-rich object insertion tasks under unstructured environments using a robot with a soft wrist, enabling safe contact interactions. For the unstructured environments, we assume that there are uncertainties in object grasp and hole pose and that the soft wrist pose cannot be directly measured.
Recent methods employ learning approaches and force/torque sensors for contact localization; however, they require data collection in the real world.
This study proposes a sim-to-real approach using a privileged training strategy.
This method has two steps. 1) The teacher policy is trained to complete the task with sensor inputs and ground truth privileged information such as the peg pose, and then 2) the student encoder is trained with data produced from teacher policy rollouts
to estimate the privileged information from sensor history.
We performed sim-to-real experiments under grasp and hole pose uncertainties. This resulted in 100\%, 95\%, and 80\% success rates for circular peg insertion with 0\textdegree, +5\textdegree, and -5\textdegree~peg misalignments, respectively, and start positions randomly shifted $\pm$ 10 mm from a default position. Also, we tested the proposed method with a square peg that was never seen during training. 
Additional simulation evaluations revealed that using the privileged strategy improved success rates compared to training with only simulated sensor data.
Our results demonstrate the advantage of using sim-to-real privileged training for soft robots, which has the potential to alleviate human engineering efforts for robotic assembly.
% , which is now a common technique in domains such as legged locomotion, to the problem setting of soft robot control.
% Sentence 1: CONTEXT - why now?
% It is now possible to capture the 3D motion of the human body on consumer hardware and to puppet in real time skeleton-based virtual characters.
% Sentence 2: NEED - why does the reader care?
% However, many characters do not have humanoid skeletons. Characters such as spiders and caterpillars do not have boned skeletons at all, and these characters have very different shapes and motions. In general, character control under arbitrary shape and motion transformations is unsolved - how might these motions be mapped?
% Sentence 3: TASK - what do we do?
% Sentence 4: OBJECT - what does this document do?
% We control characters with a method which avoids the rigging-skinning pipeline --- source and target characters do not have skeletons or rigs. We use interactively-defined sparse pose correspondences to learn a mapping between arbitrary 3D point source sequences and mesh target sequences. Then, we puppet the target character in real time.
% Sentence 5: FINDINGS - what did we discover?
% We demonstrate the versatility of our method through results on diverse virtual characters with different input motion controllers.
% Sentence 6: CONCLUSIONS - so what?
% Sentence 7: PERSPECTIVES - what now?
%Our method provides a fast, flexible, and intuitive interface for arbitrary motion mapping which provides new ways to control characters for real-time animation.

\end{abstract}

%%%%%%%%%%%%%%%%%%%%%%%%%%%%%%%%%%%%%%%%%%%%%%%%%%%%%%%%%%%%%%%%%%%%%%%%%%%%%%%%
\section{INTRODUCTION}
This study addresses the problem of controlling contact-rich manipulation, focusing on object insertion tasks under grasp and hole pose uncertainties. Enabling robots to deal with such uncertainty will significantly alleviate engineering efforts in performing elaborate calibration procedures and developing jigs, as is necessary for industrial assembly.

\begin{figure}[t]
    \centering
    \includegraphics[width=\linewidth]{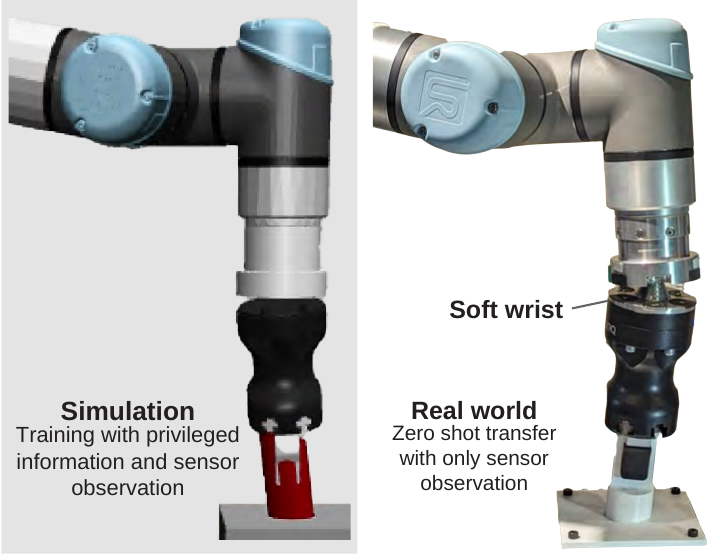}
    \caption{We propose a sim-to-real approach for object insertion for a robot with a soft wrist through a privileged training strategy.}
    \label{figure:fig-1}
\end{figure}

In the presence of such uncertainties, compliance to handle environmental contacts becomes necessary for object insertion. 
Higher compliance is desirable for more severe uncertainty.  
Many approaches adopt force control with rigid robots; however, they may struggle to achieve high compliance due to the limited bandwidth of the servo controller~\cite{ZhangRAL2023}. To address this, physically soft robots incorporating passive compliance in their mechanical design have been successfully applied to object insertion tasks~\cite{ZhangRAL2023, hartisch2023high, azulay2022haptic}. This study employs a soft wrist~\cite{von2020compact}, which provides large six Degree of Freedom (DOF) deformations (Fig.~\ref{figure:fig-1}).  
% \item Although the first passive compliant manipulators were devised in the 1980s, their range of motion was limited and therefore had limited applicability to variations in inserted object shapes
% \item However in recent years, improved designs of passive compliant manipulators have been developed which enable object insertion under high levels of uncertainty

Despite its advantages, soft robots' structural compliance presents challenges for control, exhibiting complex behavior due to its nonlinear dynamics~\cite{kim2021review}, and limitations in partial observation of the passive DOFs.  
%Thus, data-driven approaches are promising for dealing with these challenges.
Past studies often use motion capture systems to obtain the robot pose for full observation, but this requires external sensor calibration and suffers from occlusion issues~\cite{yasa2023overview}. Therefore, it is desirable for an end-use deployment to not rely on such systems.

To address these challenges, recent approaches propose reinforcement learning (RL) and object localization techniques using force/torque (FT) sensors for insertion tasks~\cite{azulay2022haptic, HaiICRA2024}. They exploit sequential FT information to explicitly or implicitly encode the relative pose between the inserted object and the hole. However, these methods require time-consuming and potentially unsafe real-world data collection, with heuristics or human demonstrations necessary. 

This study proposes a sim-to-real approach to reduce the burden of real world data collection.
We employ privileged reinforcement learning~\cite{lee2020learning}, which decomposes the learning process into two stages.
First, a teacher agent with access to privileged knowledge, including the peg pose and alignment state, learns a policy to complete the task. Then, a student agent with no privileged information imitates the teacher by estimating this information from data obtainable on the robot, including arm pose and wrist FT sensor readings.
%Privileged training is more successful than methods that directly train policies in sparse supervisory signals~\cite{lee2020learning}, like our setting. 
We develop a simulator suitable for sim-to-real transfer based on a previous study~\cite{HaiICRA2024} and compensate for remaining sim-to-real gaps through the judicious use of domain randomization~\cite{peng2018sim}.
%These techniques are employed to handle the difficulty of applying sim-to-real techniques to soft robots, arising from the tendency for policies trained purely in simulation to overfit it, which is further exacerbated by soft robot-specific challenges such as accurate physical modeling and limited state observability.

To evaluate our proposed system, we performed real robot and simulation experiments.
The real robot experiments demonstrate successful zero-shot transfer under hole and grasp pose uncertainty.
The simulation experiments reveal that privileged training improves performance compared to a similar method without this input, as well as the effectiveness of domain randomization and including peg alignment states in the privileged information.

\textbf{Contribution}: the sim-to-real transfer of contact-rich manipulation with a soft wrist using a privileged learning framework. Codes for our simulator and student-teacher training are available in the following link\footnote{\url{https://omron-sinicx.github.io/soft-robot-sim-to-real/}}. 
% Machine learning techniques offer an effective approach to generalization over object shapes
% Past works using learning-based control for passive compliant robots relied primarily on external sensors such as cameras or motion capture systems to estimate the pose of the grasped object. However, cameras suffers from issues such as occlusion, and motion capture systems are costly. Additionally, if it is desired to have an end-use deployment that does not rely on these sensors, a time-consuming pretraining phase becomes necessary, which is especially difficult if generalization over peg shapes is desired

%through this approach, we demonstrate sim-to-real peg-in-hole insertion of a variety of peg geometries, and show through ablations that the two stage privileged training framework is advantageous in the presence of passive compliance

\section{RELATED WORK}

\subsection{Soft robots for object insertion}
%The use of soft materials such as springs and elastomers in robots for object insertion has been devised as early as the 1980s~\cite{GotoSMC1980, whitney1982quasi}. Recent mechanisms can handle larger uncertainty owing to larger and multi-directional spring deformations~\cite{NishimuraHumanoids2017, von2020compact, ZhangRAL2023}.
%Designs using other modalities of softness have also been explored, such as compliant fingers driven by tendons and springs~\cite{azulay2022haptic}, fin-ray effect grippers~\cite{hartisch2023high}, and pneumatic soft grippers~\cite{brahmbhatt2023zero}.

A variety of modalities have been employed in applying softness mechanisms for robotic object insertion, ranging from springs and elastomers~\cite{ZhangRAL2023, von2020compact, GotoSMC1980, whitney1982quasi, NishimuraHumanoids2017}, compliant fingers driven by tendons and springs~\cite{azulay2022haptic}, fin-ray effect grippers~\cite{hartisch2023high}, and pneumatic soft grippers~\cite{brahmbhatt2023zero}. Learning approaches incorporating force and tactile inputs have been effective in reconciling the partial observability of soft robot control.
Azulay et al. combined reinforcement learning and heuristic force-based control in the technique that they called Haptic Glance~\cite{azulay2022haptic}.
%This supervised learning approach improves robustness by informing whether the inserted object is inside or outside the hole from FT sensor readings.
Royo-Miquel et al. used an approximate linear model to estimate the soft robot's pose via self supervised learning on tactile sensory inputs in the frequency domain~\cite{royo2023learning}.
%They developed a self-supervised approach with tactile inputs to detect when the peg has fit inside of the hole by observing tactile signal changes in the frequency domain~\cite{royo2023learning}.
Nguyen et al. employed reinforcement learning, relying on a recurrent structure and geometrical symmetries to accelerate learning~\cite{HaiICRA2024}.
%a reinforcement learning method, where the policy using a recurrent structure implicitly encodes object pose from FT sensor history, in addition to geometrical symmetries and demonstrations, to accelerate learning~\cite{HaiICRA2024}.
These methods demonstrate robust insertion, but require heuristics and data collection in the real world.
In contrast, we use a sim-to-real approach for object insertion tasks on a robot with a soft wrist.
%We train the insertion policy and pose estimation via teacher-student training purely in simulation.
Although~\cite{brahmbhatt2023zero} also used sim-to-real learning with a soft gripper robot, it did not model the compliant dynamics of the gripper in simulation.
%One study that uses sim-to-real learning with a soft gripper robot is~\cite{brahmbhatt2023zero}; however, it did not model the compliant dynamics of the gripper in simulation. 
Conversely, we develop a simulation environment explicitly modeling the compliant dynamics, leveraging MuJoCo's ability to model passive joints actuated by spring-damper systems~\cite{todorov2012mujoco}.

\subsection{Proprioception in soft robotic sim-to-real transfer}
Many studies have successfully achieved sim-to-real transfer for soft robot control~\cite{bacher2021design}. For example, accurate position tracking has been demonstrated by obtaining dynamics using simulation with Finite Element Method (FEM)~\cite{du2021underwater, dubied2022sim, zhang2022sim2real}, learning inverse kinematics~\cite{fang2022efficient}, or through vision systems~\cite{shentu2023moss, YooICRA2023}. 
Agile maneuvers of a pneumatic soft robot have also been demonstrated, but the robot pose was obtained from a motion capture system~\cite{jitosho2023reinforcement}.
Graule et al. used an approximate model using segmented rigid components for in-hand manipulation with a pneumatic hand~\cite{graule2022somogym}.
An Extended Kalman Filter was used for joint angle estimation for a tendon-driven hand~\cite{toshimitsu2023getting}.
Unlike these studies, we propose a student-teacher training approach to simultaneously obtain proprioceptive state estimation and control policies.

\subsection{Privileged training for robot control}
Privileged training was initially proposed for autonomous driving~\cite{learning-by-cheating} but has since been used for various robot applications. This includes quadrupedal locomotion~\cite{lee2020learning, kumar2021rma}, aerial robots~\cite{loquercio2021learning}, cloth manipulation~\cite{lin2022learning}, push manipulation~\cite{kim2023pre}, and in-hand manipulation~\cite{qi2023general, rostel2023estimator}. 
Our study is the first attempt to use privileged training for a soft robot's object insertion task. 
%While most works used environmental conditions and target object poses as privileged information, we propose to use this technique for estimating the internal states of a soft robot and task completion progress.
In contrast to previous studies using privileged RL primarily to predict information relating to interaction states and parameters external to the robot, such as ground contact, terrain height, or grasped object properties~\cite{lee2020learning, kumar2021rma, qi2023general}, our use of it can be interpreted as a method of producing an estimation module for states internal to a soft robot which cannot easily be observed directly without additional sensors.
\section{PRIVILEGED TRAINING} \label{section:method}
\subsection{Problem Statement} \label{section:method-problem-statement}

% \leavevmode
% \begin{wrapfigure}[15]{R}{0.5\linewidth}
%   \centering
%   \includegraphics[width=\linewidth]{figures/problem-v3.pdf}
%   \caption{The problem setting considered in this work, showing variable definitions for quantities observable from sensors, versus privileged information only accessible in simulation.}
%   \label{fig:coordinates}
% \end{wrapfigure}

\begin{figure}
    \centering
    \includegraphics[width=0.7\linewidth]{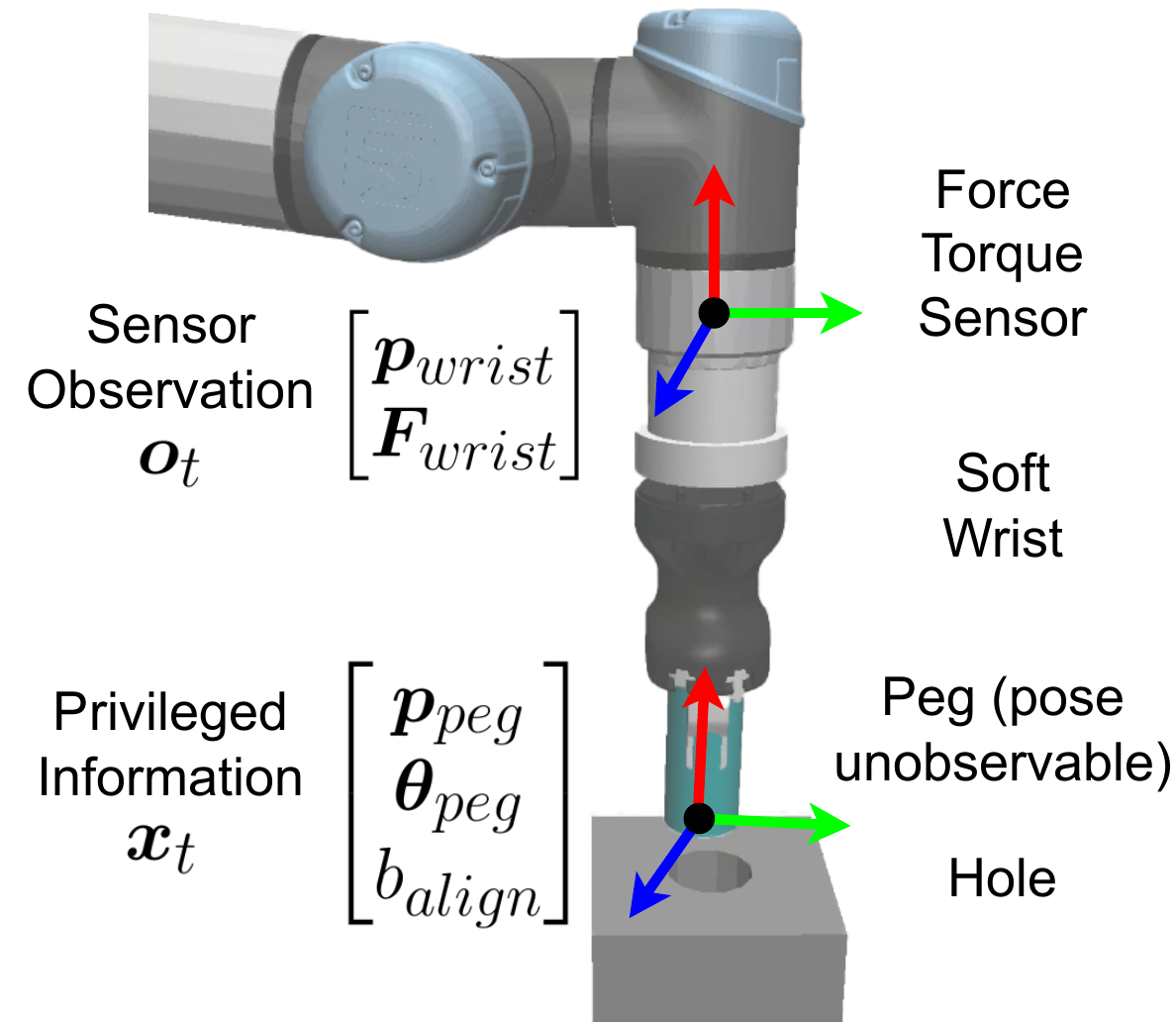}
    \caption{The problem setting considered in this study, showing variable definitions for quantities observable from sensors, versus privileged information only accessible in simulation.}
    \label{fig:coordinates}
\end{figure}

\begin{figure*}[t]
    \centering
    \includegraphics[width=\linewidth]{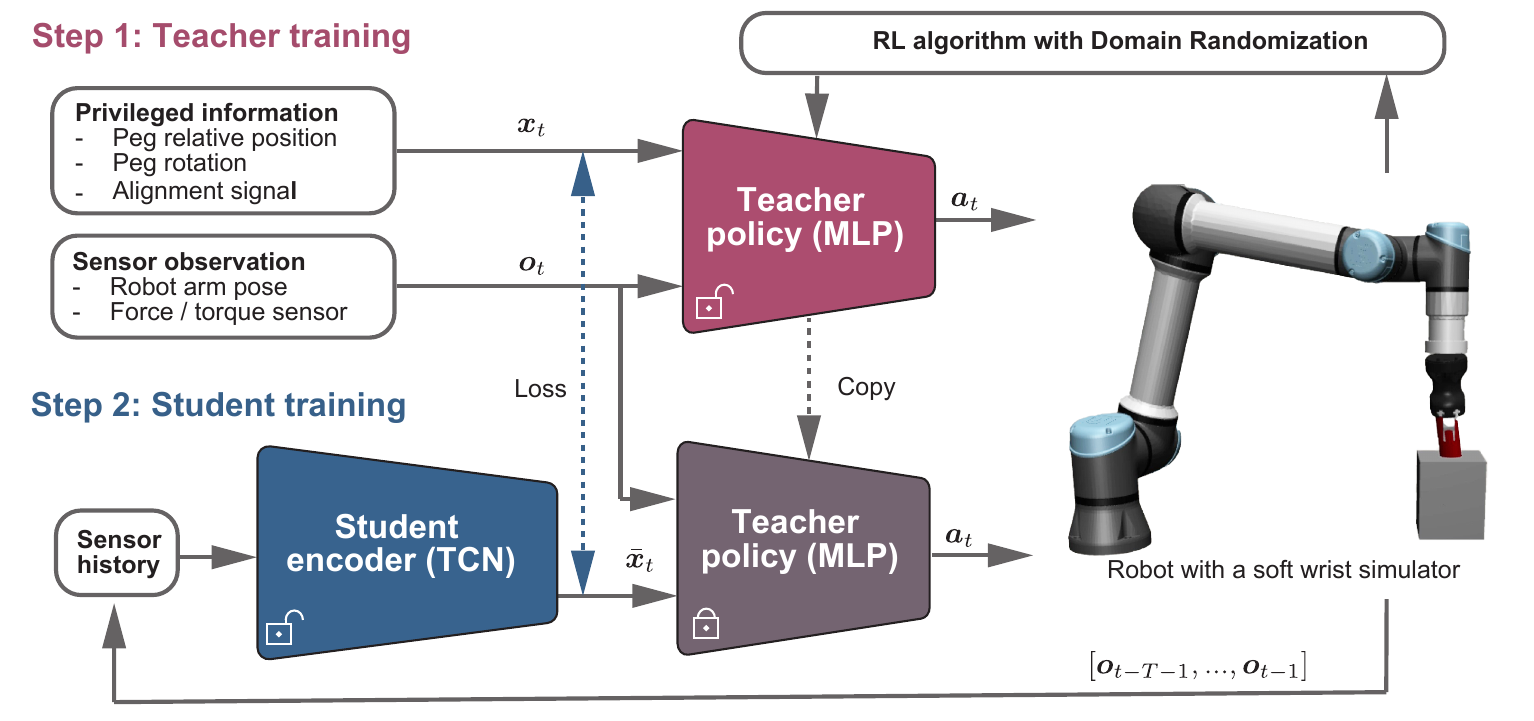}
    \caption{Overview of the proposed framework. The privileged training has two phases: 1) teacher training: the teacher policy to control the robot is trained with sensor inputs and ground truth privileged information, and 2) student training: the student encoder is trained by running the learned teacher policy to estimate the privileged information from sensor history.}
    \label{fig:overview}
\end{figure*}

The problem setting in this study is illustrated in Fig.~\ref{fig:coordinates}.
The Robotiq Hand-E gripper is attached to the Universal Robotics UR5e robot arm through the soft wrist connection introduced in~\cite{von2020compact} consisting of three springs. The gripper grasps peg-like objects, and the objective is to insert the grasped object into a hole. The controller has access to sensory data on the robot, including the wrist pose and readings from a 6-axis FT sensor attached to the end effector above the soft wrist.

The passive compliance in the soft wrist enables safe contact interactions suitable for this peg-in-hole task~\cite{von2020compact}. However, it also introduces a challenge for control since the gripper pose cannot be observed directly without external sensing hardware such as cameras or motion capture devices. Moreover, we assume that the in-grasp pose of the peg is not precisely known, and there is uncertainty about the hole location relative to some known nominal location. We introduce these sources of uncertainties and opt not to use external sensing hardware in light of the motivation of achieving industrial assembly tasks without expensive calibration effort and hardware necessary to remove such uncertainties. In this setting, the goal is to predict the pose of the peg and carry out the insertion task with the limited sensor data available on the robot.

\subsection{Overview}
The proposed system is shown in Fig.~\ref{fig:overview}. The inability to directly observe certain system states is solved through privileged RL, which is a sim-to-real model-free RL method based on~\cite{learning-by-cheating}. This approach has two training phases. In the first phase, the teacher policy network is trained using privileged information only accessible in simulation. In the second phase, a student network is trained to enable imitation of the teacher's motions with access only to simulated sensor data. The result can then be transferred directly to real hardware since the necessity to rely on privileged information is removed through the second training phase. 
%For both the teacher and student training, we use a modified version of the Robosuite~\cite{robosuite2020}, which uses MuJoCo~\cite{todorov2012mujoco} as the underlying physics simulator. We describe each component in further detail in the following sections.

\subsection{Privileged Teacher Training}
In the first phase of teacher training, a policy is trained to complete the peg-in-hole insertion task with access to simulator information not accessible on the real robot. We do this by formulating the problem as a Markov Decision Process (MDP) and solve it using Proximal Policy Optimization (PPO)~\cite{schulman2017proximal} with simple 2-layer $256 \times 256$ Multi-Layer Perceptron (MLP) actor and critic networks. The definitions of each component of the MDP are outlined below.

\subsubsection{State} \label{section:method-observation}
We assume that the environment is fully observable during teacher training. In addition to the end effector pose and wrenches obtained through simulated sensors, the teacher policy has privileged access to the pose of the peg and its alignment state within the hole. The alignment state can be interpreted as a subtask switching condition facilitating more robust task completion.
In practice, we remove the rotational components of the pose and wrench data, resulting in an $\mathbb R^6$ vector of wrist position $\bm p_{wrist}$ and force $\bm F_{wrist}$ as the simulated sensor observation $\bm o_t$. This was because commanding translational motions was found to be sufficient given the rotational degrees of freedom provided passively to the peg through the flexible wrist, and we also observed that torque readings have negligible magnitude compared to forces given the mechanical compliance structure.

For the privileged information $\bm x_t \in \mathbb R^{10}$, we use the position $\bm p_{peg}$, orientation $\bm \theta_{peg}$, and binary alignment state $b_{align}$ of the peg, which returns true when the pose error between the peg and hole is smaller than a threshold. The peg orientation is represented as an $\mathbb R^6$ vector consisting of the first two columns of the $SO(3)$ rotation matrix, as this representation was shown to be effective for neural network regression in~\cite{zhou2019continuity}, which will be necessary for the student training phase.
All observations are normalized with manually specified offset and scale factors to make values roughly lie within $[-1,1]$ during an insertion motion.
%, and the alignment state is mapped to 1 or 0 for true and false, respectively.
%Identical scale and offset factors are applied to the physical robot during deployment.

\subsubsection{Action} \label{section:method-action}
As described in the previous section, linear translations of the wrist were found to be sufficient to enable peg insertion without the need for rotations. In particular, we define actions for our MDP as end effector position changes relative to the current position, $\bm a_t = \Delta (\bm p_{wrist})_{t}$, with a maximum value of 3\si{\milli \meter} over a 50\si{\milli \second} interval corresponding to the 20\si{\hertz} policy frequency, for each of the three coordinate axes. These position commands are converted to linear trajectories through Robosuite's built-in interpolator function, so they can be tracked smoothly using joint torques calculated with the built-in implementation of Operational Space Control (OSC)~\cite{khatib1987unified} at the MuJoCo default simulation frequency of 500\si{\hertz}. We found that the default OSC parameters in Robosuite were overly compliant for accurate pose tracking and were not reflective of motions on the physical robot, so we increased and randomized it as discussed in Sec.~\ref{section:method-randomization}.

\subsubsection{Reward} \label{section:method-reward}
The reward function used in this study is inspired by~\cite{hamaya2021robotic} and has the form \useshortskip
\begin{align}
    r = r_p - r_i - r_a + r_s,
\end{align}
where
\begin{enumerate}
     \item[(1.a)] $r_p = (d_{t-1} - d_t) / 0.001$ is the progress reward, where $d_t \in \mathbb R$ is the weighted peg distance at timestep $t$ defined as $d_t = \sqrt{\bm e_t^T \bm W \bm e_t}$, with $\bm e_t \in \mathbb R^3$ representing the positional error between the peg tip and the hole, and $\bm W \in \mathbb R^{3 \times 3}$ being a diagonal weight matrix with elements $[1, 1, 10]$.
    \item[(1.b)] $r_i = w_i (a_t)_z^2$ is the insertion reward, where $(a_t)_z$ is the action along the vertical axis, and $w_i$ is $0.001$ if the alignment condition defined below is satisfied, and $1$ otherwise. 
    \item[(1.c)] $r_a = ||\bm a_{t} - \bm a_{t-1}||^2$ is the action smoothness reward designed to prevent the policy from producing vibrations and jerky motions that are not conducive to safe and successful sim-to-real transfer.
    \item[(1.d)] $r_s$ is the sparse success reward having value 1 if insertion is successful as defined by $||\bm e_t|| < 0.005$, and 0 otherwise.
\end{enumerate}

\subsubsection{Alignment}\label{section:method-alignment}
The binary alignment condition is defined as the state where translational peg position errors are within some threshold, \useshortskip
\begin{align}
    b_{align} = \sqrt{(e_t)_x^2 + (e_t)_y^2} < 0.007.
\end{align}
This is used both for defining weights in the insertion reward in Sec.~\ref{section:method-reward}, and as a privileged observation in Sec.~\ref{section:method-observation}. Modulating the vertical action weight according to peg alignment was determined in~\cite{hamaya2021robotic} to be crucial for the soft wrist peg-in-hole task, as the robot should drag the peg lightly on the flat surface adjacent to the hole until it senses through contact interactions and FT sensor readings that the hole is found and the peg is vertically aligned. Allowing excessive vertical force before alignment results in policies converging to local minima, where friction prevents the peg from moving toward the hole after contact is made.
%This conditional reward weight switching can be interpreted similarly as a switching condition for a staged reward, corresponding to different subtasks of the peg-in-hole insertion task~\cite{lee2019making}. Unlike~\cite{lee2019making}, however, we found that only one switching condition was necessary, and the remaining subtasks, such as reaching and insertion, emerged without explicit human specification.

\subsubsection{Termination}
Additionally to the rewards, early episode termination can be used to specify desired motions, to avoid undesired local minima and to prevent the agent from collecting data in state space regions irrelevant to the task~\cite{peng2018deepmimic}. In this study, we terminate the episode and add an additional $-5$ penalty to the reward whenever the weighted peg distance $d_t$ becomes 1.2 times the value at the beginning of the episode, typically indicating a local minimum where the peg is placed at the correct height outside of the hole without attempting insertion. The same penalty is applied if the maximum allowable episode length of 200 timesteps, corresponding to 10 seconds, is reached without successful insertion. We additionally terminate the episode without penalty after successful insertion.
%, as we wish to collect experience data for the motion leading up to insertion rather than afterward.

% The initial pose of the robot is also randomized by adding Gaussian noise to all six joints with a standard deviation of 0.002~\si{\radian}. This is with respect to a configuration where the peg nearly touches the flat surface adjacent to the hole. All randomizations are applied at the beginning of every episode, and unless otherwise noted, they are uniformly distributed.

\subsection{Student Training} \label{section:method-student}
After the RL-based teacher policy training has converged, the second phase consists of training a student encoder network to imitate teacher policy behaviors despite using only wrist position and force sensory inputs as per Sec.~\ref{section:method-observation}. Specifically, we define a student encoder network to take a horizon $T=20$ sensor history $[\bm o_{t-20}, \ldots \bm o_{t-1}]$ corresponding to 1 second of sensor reading data as inputs, and outputs an estimate $\bar {\bm x}_t$ of the privileged data defined in Sec.~\ref{section:method-observation}. This history buffer is initialized with zeros at the beginning of the episode. A Temporal Convolutional Network (TCN)~\cite{bai2018empirical} was used as the architecture similarly to~\cite{lee2020learning}. As per~\cite{lee2020learning}, we train this using supervised learning with data collected through student rollouts to avoid distribution shift issues between the teacher and student~\cite{ross2011reduction}. The loss between predicted and ground truth values is given as: \useshortskip
\begin{align}
    \mathcal{L}(\bar{\bm{x}}, \bm{x}) = \mathcal{L}_{mse}(\bar{\bm{x}}_p, \bm{x}_p) + w\mathcal{L}_{bce}(\bar{b}_{align}, b_{align}), 
\end{align}
where $\bm{x}_p$ is composed of the peg pose $\bm{p}_{peg}$ and orientation $\bm{\theta}_{peg}$, $\mathcal{L}_{mse}$ is the mean square error loss, $\mathcal{L}_{bce}$ is the binary cross entropy loss, and $w$ is a weight set to 0.1.
Unlike~\cite{lee2020learning} and more similarly to~\cite{kumar2021rma}, we do not retrain the policy and only train the encoder network in the second phase. Unlike both~\cite{lee2020learning} and~\cite{kumar2021rma}, however, we do not use a latent embedding of estimated variables and instead directly regress on physical quantities. In addition to providing outputs that are human-interpretable and simple to implement, we show through experiments that this was sufficient in our setting to provide accurate estimates.

\section{SIM-TO-REAL TRANSFER}

\subsection{Soft wrist simulator}
In order to adapt Robosuite~\cite{robosuite2020} to model our problem setting, we created custom MuJoCo XML files for the flexible wrist and gripper assemblies. We converted the Robotiq Hand-E URDF file provided in~\cite{beltran2024robotiq} to a MuJoCo XML file and rigidly attached peg objects to it under the assumption that no in-grasp slip occurs, and did not model finger actuation. We model the flexible wrist mechanism as a series of four springs, each providing one passive DOF along four axes parametrized by $k = [k_z, \kappa_x, \kappa_y, \kappa_z]=[1000, 0.5, 0.5, 5]$ and $b = [b_z, \beta_x, \beta_y, \beta_z] = [1, 0.005, 0.005, 1]$, where $k, \kappa$ are stiffness constants and $b, \beta$ are damping constants along linear and rotational axes respectively, and SI units used for all values. The simplification of not modeling the x-y translational DOFs of the physical 6 DOF flexible wrist was made under the observation that x-y forces applied on the peg tip primarily caused tilting along y-x axes respectively, with negligible translational motions.

Given that the spring parameter values were determined by making simulated wrist motions visually similar to the physical robot's, a more accurate system identification procedure could likely be performed through calibration involving curve fitting to motion capture data. However, we did not perform this given the project motivation described in Sec.~\ref{section:method-problem-statement}, and instead show that successful control can be achieved despite highly approximate setup procedures. Following~\cite{HaiICRA2024}, we decompose hole object geometries into convex meshes to enable contact simulation within MuJoCo.

\subsection{Domain randomization} \label{section:method-randomization}
During teacher and student training, we apply domain randomization, a common technique in sim-to-real RL to enable robust transfer of policies to physical hardware~\cite{peng2018sim}. In this study, we randomize the in-grasp angle of the peg within $\pm$5\si{\degree} and the position of the hole within $\pm$10\si{\milli \meter}.

While the above randomization is designed to enable practical task generalization, we additionally randomize gains in the OSC controller as mentioned in Sec.~\ref{section:method-action}. Position controller gains are randomized uniformly on a logarithmic scale between $[10^3, 10^4]$ with the same values used for all 6 DOF. 
%This is in comparison to the default Robosuite values of 150 and 1.0, respectively. 
This was done to account for the discrepancy that the simulation and OSC formulation assume a torque-controlled robot, whereas the physical UR5e has mechanically stiff joints and precise position control implemented at a low level. 
While one could perform system identification to determine the simulated OSC parameters that best correspond to the physical robot motions, we instead opted for randomization, as we expect generally robust policies to emerge due to the high sensitivity of motions to gain parameters that the policy must learn to adapt to.

The initial pose of the robot is also randomized by adding Gaussian noise to all six joints with a standard deviation of 0.002~\si{\radian}. This is with respect to a configuration where the peg nearly touches the flat surface adjacent to the hole. All randomizations are applied at the beginning of every episode, and unless otherwise noted, they are uniformly distributed.

\subsection{Real robot controller}
In order to execute trained networks to control the physical robot, the code in~\cite{beltran2024ur3} was used, which provides Python interfaces to Universal Robot's ROS drivers~\cite{ur-ros-driver}, providing convenient interfacing with our PyTorch~\cite{paszke2017automatic} based training pipeline. Important features include the ability to specify the final position and time of end effector trajectories and the ability to interrupt commands to ensure a 20Hz update frequency in case a path execution takes longer than the specified time. Given the mismatch of compliant OSC-based control in simulation and high precision position control on the physical robot as discussed in Sec.~\ref{section:method-randomization}, ensuring that position displacement targets were tracked accurately and at the same frequency for both simulated and real robots was found to be crucial for sim-to-real transfer.
\section{RESULTS}
\subsection{Overview}
We perform a series of experiments on real and simulated robots to characterize the performance of the proposed method, as well as the effect of various design parameters on performance. In particular, we 1) show the overall sim-to-real performance and report success rates on the real robot, 2) demonstrate the advantage of using the two-stage privileged RL approach, 3) evaluate the importance of domain randomization, and 4) visually show the effectiveness of estimates produced by our student encoder. 

For training, the teacher policy and student encoder were updated with 10,000 and 5,000 iterations, respectively, with each iteration consisting of 1,000 timesteps. The teacher and student training took approximately 24 and 12 hours, respectively, on a workstation with an AMD Ryzen Threadripper Pro 5975WX CPU and an RTX 4080 GPU. 

A circular peg with 40~\si{\milli \meter} outer diameter and a hole with 42~\si{\milli \meter} inner diameter was used to train the policy for all experiments, with similarly dimensioned pegs used for tests involving other peg shapes. In the real robot experiments, we used 3D-printed pegs and holes.

\subsection{Sim-to-Real Peg-in-Hole Insertion}
\begin{figure}
    \centering
    \includegraphics[width=\linewidth]{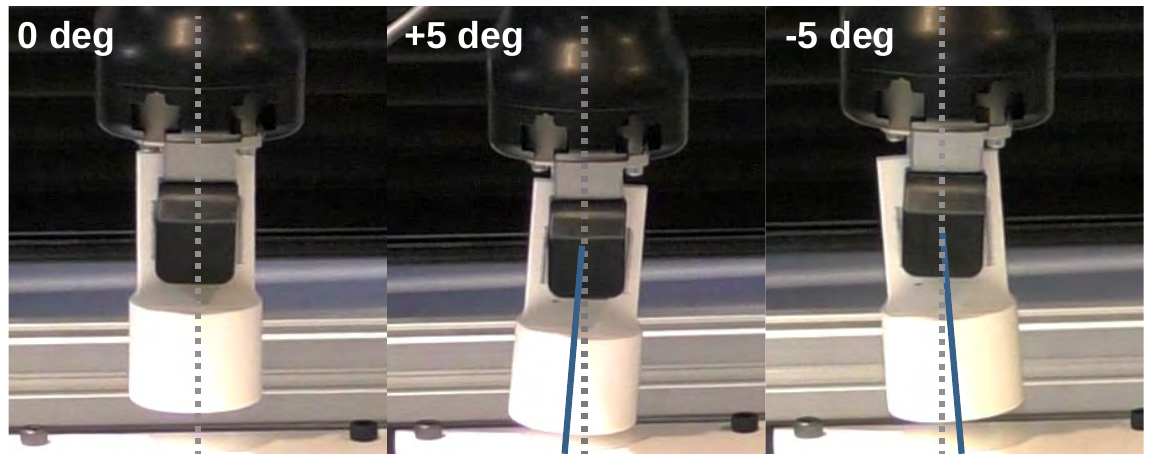}
    \caption{Pegs' misalignments for the real-robot experiments.}
    \label{fig:misalignment}
\end{figure}

Table~\ref{tab:success} shows the number of successes of the real-robot experiments with circle and square pegs under uncertainty, where 0 and $\pm$5~\si{\degree} misalignments were provided (Fig.~\ref{fig:misalignment}), and the initial x and y positions were uniform-randomly shifted in a range $\pm$10~\si{\milli \meter} from a default position. We performed 20 trials on each condition. We used teacher and student networks trained on the circular peg for both insertion tasks, meaning that the robot never saw the square peg during training. The experiment was performed by selecting the best and median-performing teacher-student network pair as evaluated in simulation. This variation across random seeds is illustrated in Fig.~\ref{fig:student} and elaborated in the next section.

The robot successfully completed circular peg insertion in most trials, despite misalignments.
While the best performing network showed a 70 \% success rate for the unseen square peg insertion without misalignment, the median-performing network showed a 45 \% success rate.
%The success rates decreased in the square insertion with the misalignments. However, in -5~\si{\degree} misalignment, the median-performing network achieved an 85 \% success rate because the generated actions from the policy and misalignment prevented the peg from sticking to the hole's surface. 
Both median and best performance decreased for the square peg in the presence of misalignment. However, the median policy actually outperformed the best policy in the -5~\si{\degree} misalignment case, as the specific motion learned by the median policy was well suited to insertion with negative angle peg insertions.

% \begin{table}
% \centering
% \caption{Number of successes of real robot experiments}
% \label{tab:success}
% \resizebox{0.7\columnwidth}{!}{%
% \begin{tabular}{ccc}
% \hline
% \multicolumn{1}{l}{} & \multicolumn{2}{c}{peg shape} \\ \cline{2-3} 
% peg misalignment         & circle      & square      \\ \hline
% 0 deg                & 20/20       & 14/20       \\
% +5 deg               & 19/20       & 9/20        \\
% -5 deg               & 16/20       & 11/20       \\ \hline
% \end{tabular}%
% }
% \end{table}

% Please add the following required packages to your document preamble:
% \usepackage{graphicx}
% Please add the following required packages to your document preamble:
% \usepackage{graphicx}
\begin{table}
\centering
\caption{Number of successes of real robot experiments}
\label{tab:success}
\resizebox{\linewidth}{!}{%
\begin{tabular}{cclcl}
\hline
\multicolumn{1}{l}{} & \multicolumn{4}{c}{Shape}                                                                   \\ \hline
                     & \multicolumn{2}{c}{Circle}        & \multicolumn{2}{c}{Square (Unseen)}                              \\ \hline
Misalignment         & \multicolumn{1}{l}{Best} & Median & \multicolumn{1}{l}{Best}   & Median                     \\ \hline
0 deg                & 20/20                    & 20/20  & 14/20                      & 9/20                       \\
+5 deg               & 19/20                    & 16/20  & 9/20                       & 2/20                       \\  
-5 deg               & 16/20                    & 18/20  & 11/20                      & 17/20 
\\ \hline
\end{tabular}%
}
\end{table}

\begin{figure}[t]
    \centering
    \includegraphics[width=\linewidth]{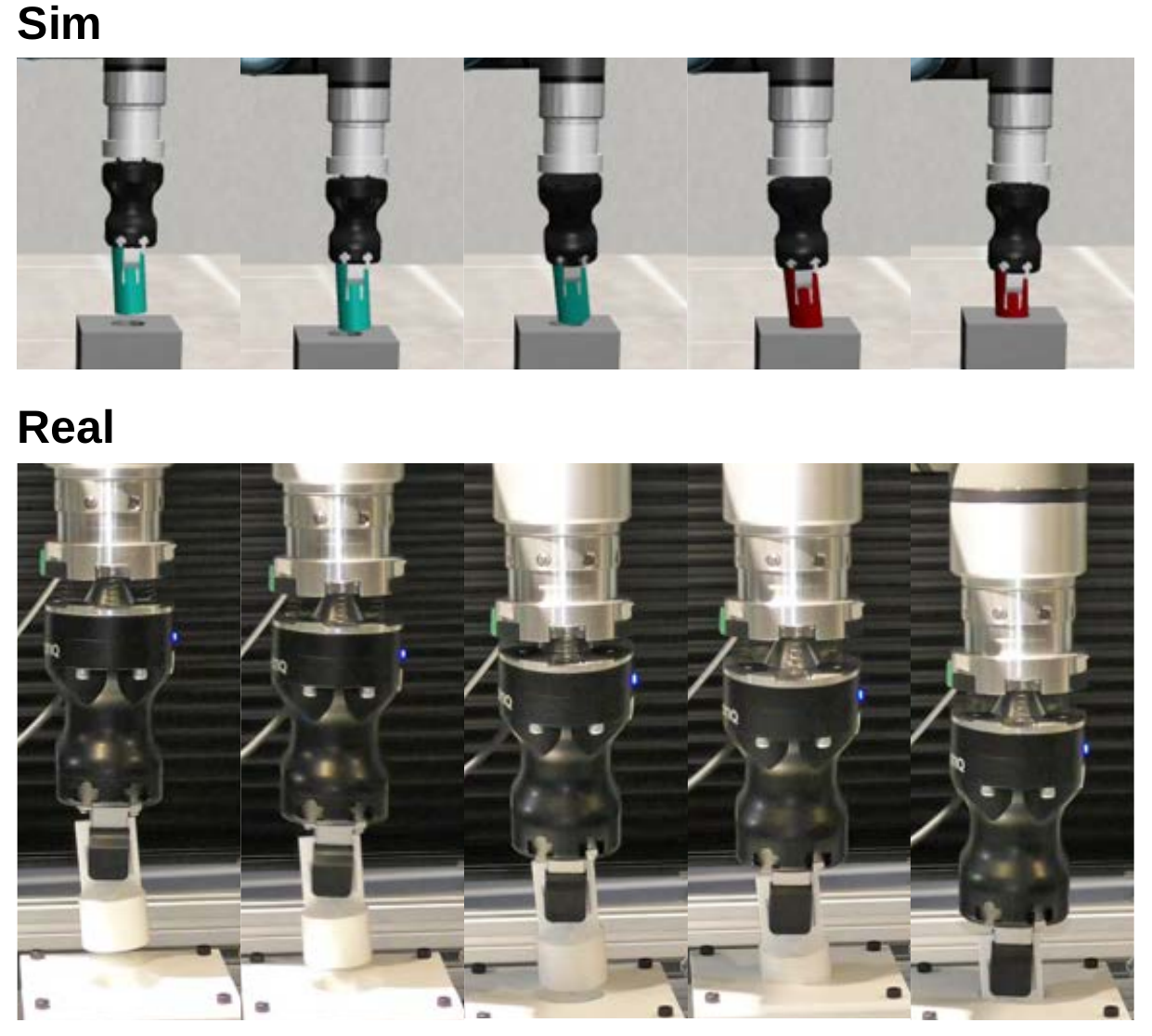}
    \caption{Snapshots of successful insertion in simulation and real-world. For the simulation, the peg's color turning red indicates that the student encoder detected the alignment of the peg.}
    \label{fig:snapshots}
\end{figure}

The attached video, as well as the snapshots provided in 
Fig.~\ref{fig:snapshots}, demonstrate zero-shot sim-to-real transfer of the contact-rich task of peg-in-hole insertion despite various sources of uncertainties and limited sensor data. In Fig.~\ref{fig:snapshots}, the color of the peg additionally provides visualization of the alignment state described in Sec.~\ref{section:method-alignment}.

\subsection{Privileged Training}
\label{section:privileged_trainnig}
\begin{figure}[t]
    \centering
    \includegraphics[width=\linewidth]{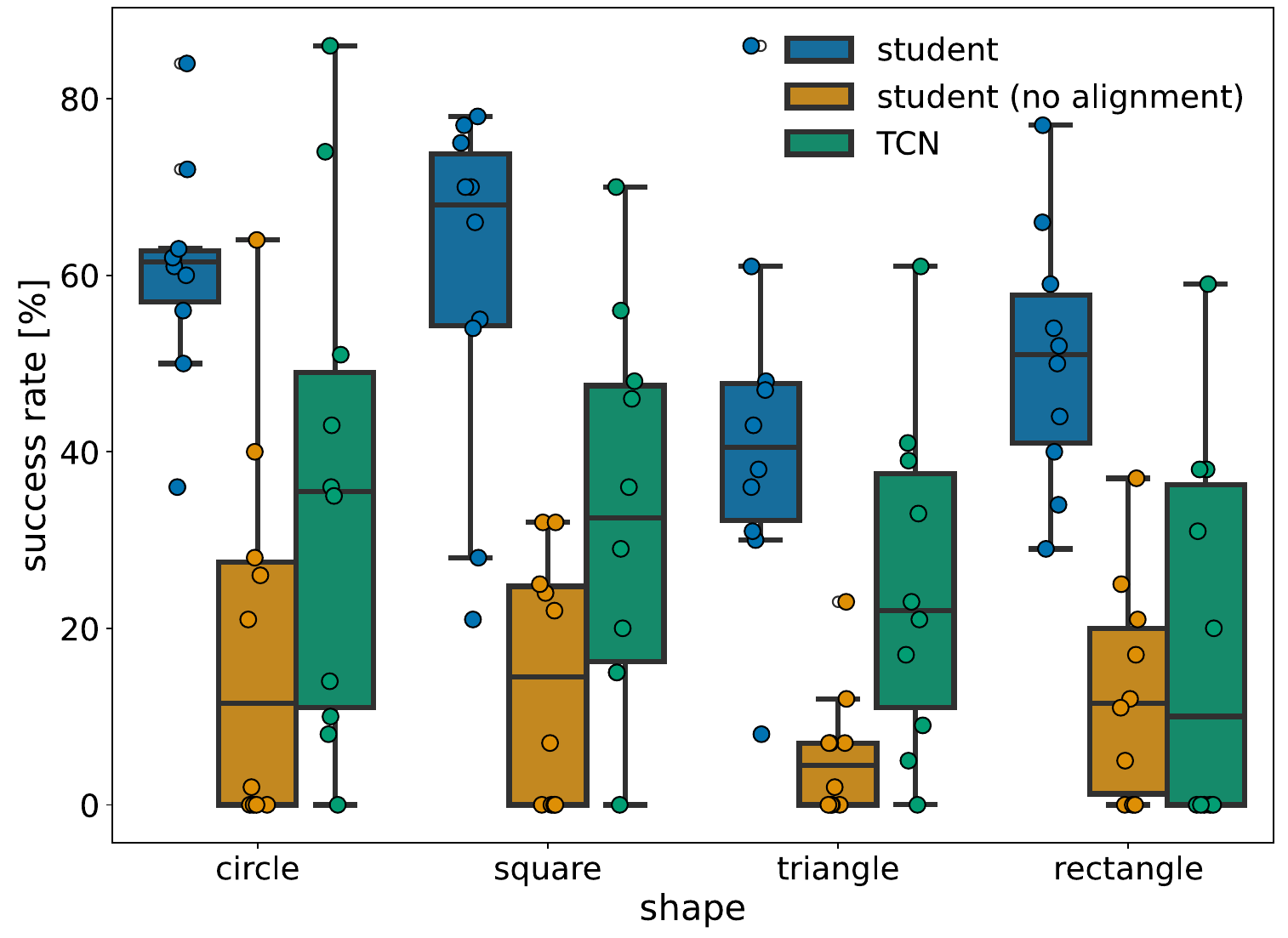}
    \caption{Success rates between the proposed method (teacher-student) and the baseline (TCN) in simulation. The scatter points show the success rates of each seed, with box plots providing summary statistics. Our proposed method, including the alignment state, showed a higher success rate than the baseline. The policy generalizes to shapes unseen during training, since only circular pegs were used for training.}
    \label{fig:student}
\end{figure}

Fig.~\ref{fig:student} shows the success rate of simulated peg-in-hole experiments to evaluate the effectiveness of the two-stage privileged training setup.
``student" denotes our proposed method, whereas ``TCN" shows a similar baseline used in~\cite{lee2020learning} where a single RL training stage is used only on simulated sensor data without privileged data, with a TCN architecture used for both actor and critic networks for a fair comparison with the two-stage approach. Additionally, ``no alignment" shows the case where peg alignment was removed from the privileged data in our proposed method. 
We trained these methods with 10 random seeds and evaluated them for 100 trials by uniformly randomizing the dynamics parameters, hole, and grasp poses in the same range as the training. 
Results are shown for four types of peg shapes, using policies trained only for the circular peg.

The results show the clear advantage of using two-stage privileged training over direct policy training with a TCN, as well as the importance of including alignment along with peg pose in the privileged data. The generalization across unseen peg shapes shows the extrapolation capabilities of RL-based controller synthesis. Despite the unexpected result that the circular peg's median success rate is outperformed by the square peg, which was unseen during training, the success rate of the square peg has a larger variance.
% This suggests that although some seeds perform well on the square peg, the performance is more consistent with the circular peg that it was trained for.
%However, the simulated success of four peg shapes serves also to expose sim-to-real gaps that remain in our system, as only circular and square pegs worked reliably in the hardware experiments described in the previous section.
In Fig.~\ref{fig:student}, it can be observed that the simulated success rates are generally lower than that of the physical experiment results provided by Table \ref{tab:success}. This is likely because the simulated training environment is harsher than the physical testing environment, since the joint controller gains are randomized during training in an effort to make sim-to-real robust, even though the joint stiffnesses are likely fixed on the physical robot.

\begin{figure}
    \centering
    \includegraphics[width=\linewidth]{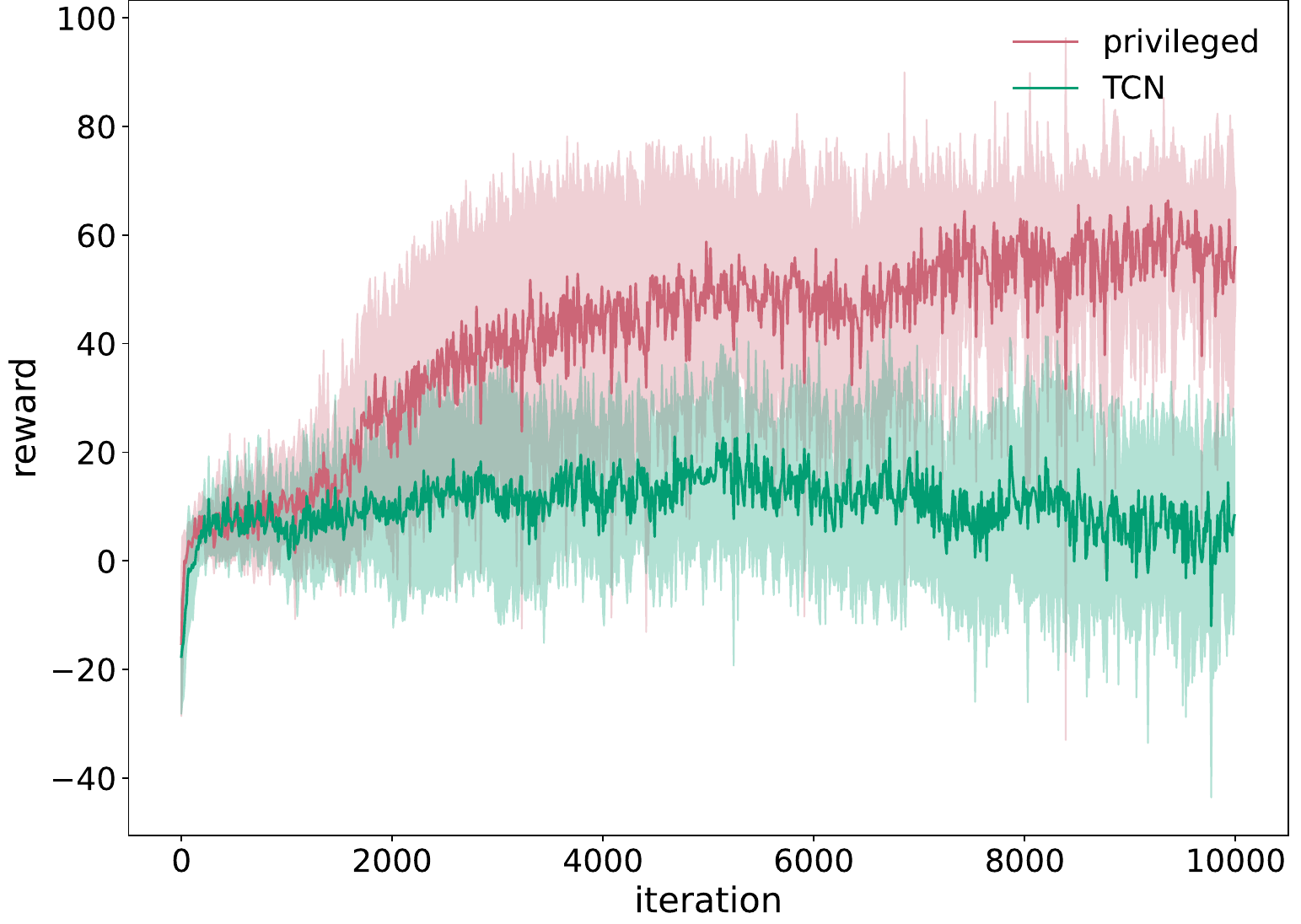}
    \caption{The learning curve of the MLP policy with privileged information and TCN policy without it. The privileged information helped the learning process.}
    \label{fig:learngcurve}
\end{figure}

In addition to evaluating success rates on converged policies, Fig.~\ref{fig:learngcurve} shows learning curves for the privileged teacher training compared against the unprivileged TCN policy baseline. The thick line and shaded regions show the mean and standard deviation of training progress across 10 experiments, each with different random seeds. This plot shows that the privileged data helps RL training despite the potential architectural advantage that the TCN policy may have over the far smaller MLP used for the privileged policy.

\subsection{Domain Randomization}
\label{section:results-domain-randomization}
\begin{figure}
    \centering
    \includegraphics[width=\linewidth]{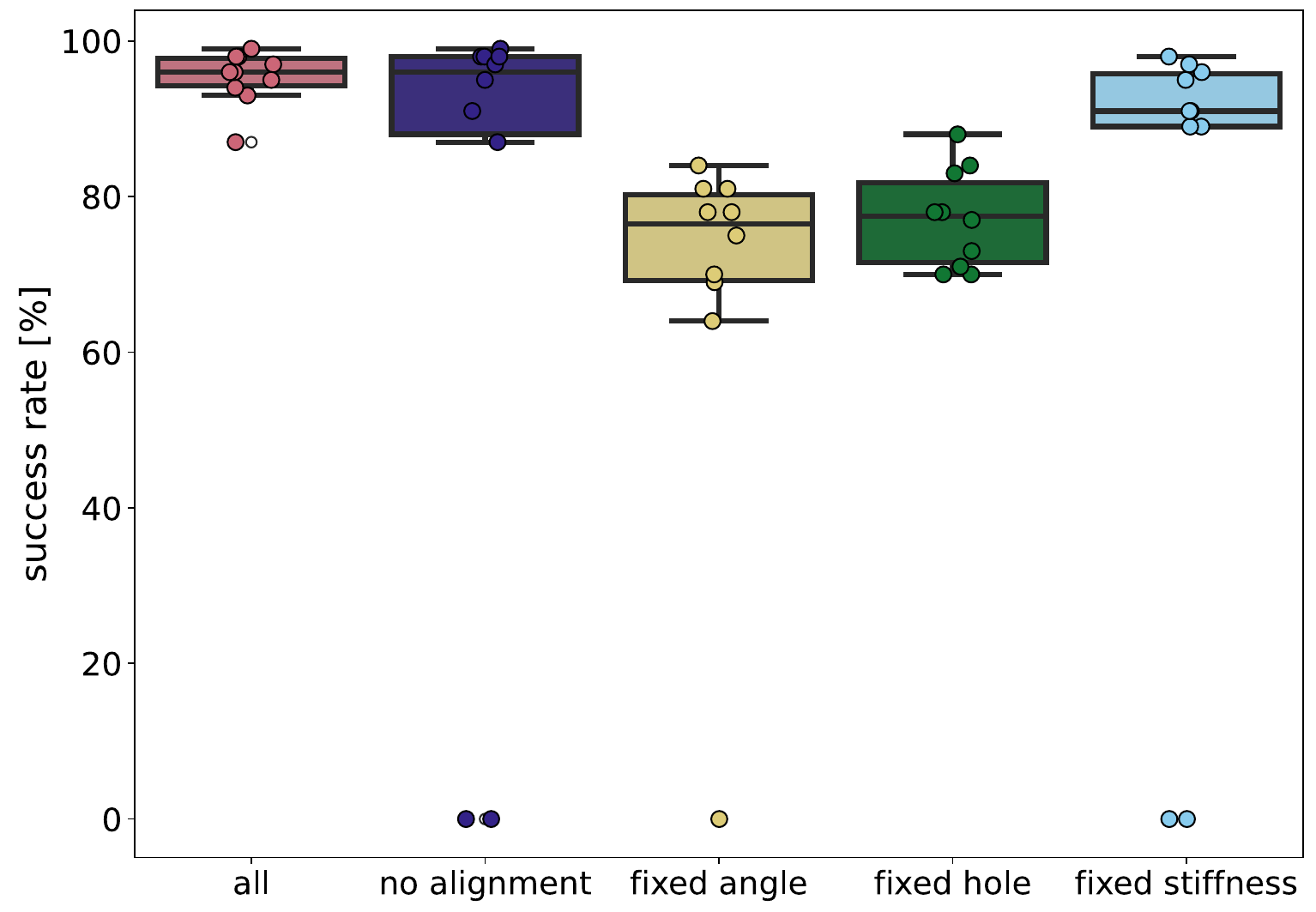}
    \caption{Success rates of the teacher policy in simulation, removing the alignment state from the privileged information and turning the randomization off during training. The success rate is degraded without the hole and grasp pose randomization.}
    \label{fig:dr}
\end{figure}

Next, we perform a sim-to-sim evaluation of domain randomization. For this, we train the privileged teacher on a simulated environment with one of the randomizations turned off and measure success rates for the same privileged policy tested on the randomized environment used in our proposed method to determine how well the policy adapts to variations unseen during training (Fig.~\ref{fig:dr}). The number of evaluation trials was the same as the previous section~\ref{section:privileged_trainnig}. 
``all" has all randomizations turned on, representing the case where the policy is tested on the same domain that it is trained on, ``fixed angle," ``fixed hole", and ``fixed stiffness" respectively represent cases where the in-grasp peg angle, hole position deviation, and OSC gain parameter randomizations are turned off during training. The results show that, especially for uncertainties in grasp and hole pose, domain randomization during training helps significantly to adapt to uncertainties that may exist during deployment.

Although unrelated to domain randomization, the ``no alignment" column in Fig.~\ref{fig:dr} shows the privileged teacher policy performance with the peg alignment state removed from the privileged information. The fact that the performance is not degraded significantly compared to the ``all" case, despite the large performance degradation observed for the student performance in Fig.~\ref{fig:student}, suggests that including the peg alignment state becomes advantageous only after the second phase of privileged training. A possible explanation is that the student encoder network can better predict the alignment state compared to other quantities, providing useful inputs to the teacher policy when privileged data is estimated imperfectly by the student encoder, as opposed to when ground truth information is provided.

\begin{figure}
    \centering
    \includegraphics[width=\linewidth]{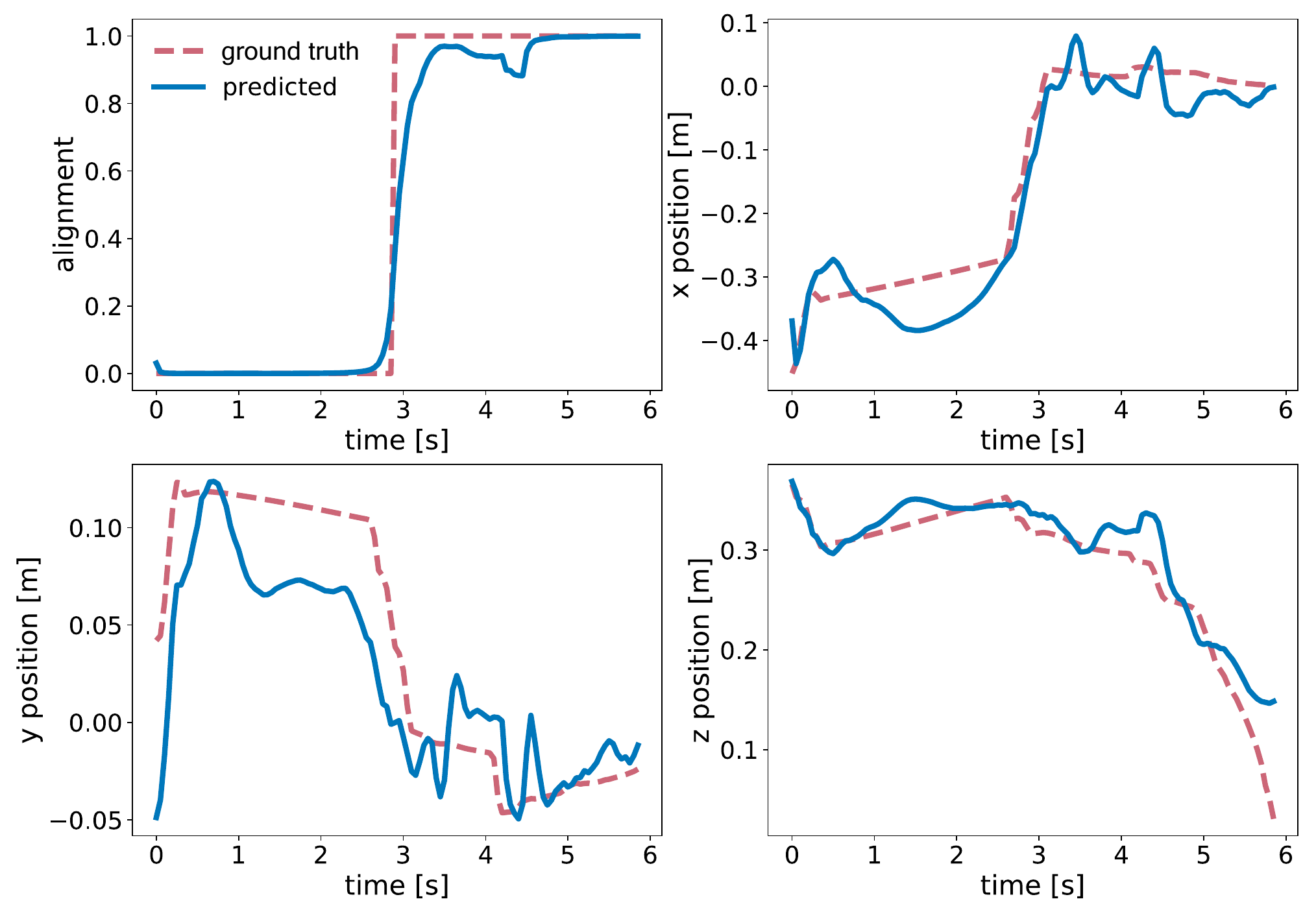}
    \caption{Student encoder predictions for the alignment state, x, y, and z positions. The encoder was successful in detecting the peg's alignment, which is defined in Sec.~\ref{section:method-alignment}.}
    \label{fig:student_prediction}
\end{figure}

\subsection{Student Network Estimation}
It was hypothesized in the previous section that the alignment state can be better predicted by the student encoder compared to other physical quantities. This is confirmed by Fig.~\ref{fig:student_prediction}, which shows the physical quantities predicted by the student encoder compared to ground truth values, as evaluated in simulation. These plots illustrate the generally high quality of the estimation. The ability to produce these plots directly without having to train a separate decoder network purely for visualization, as was done in~\cite{lee2020learning}, is useful for debugging purposes and demonstrates the advantage of performing regression directly on physical quantities rather than on latent embeddings, as discussed in Sec.~\ref{section:method-student}.

\section{CONCLUSION AND FUTURE WORK} \label{section:conclusion}
In this study, we present a method for controlling peg-in-hole insertion for an industrial manipulator robot with a soft wrist and uncertainties in physical setup, relying only on commonly available sensors without additional sensing hardware or calibration procedures. We do this through zero-shot sim-to-real transfer of neural networks trained through a privileged learning approach, leveraging the fact that training happens purely in simulation to use privileged data unavailable on the physical system. 

% Despite techniques proposed to combat this, sim-to-real gaps remain in our system.
%, as evident through the success of all four peg geometries in simulation despite only circle and square pegs achieving reliable insertion on hardware. 
A common failure mode was the peg ``missing" the hole and never ``finding" it through contact, as the policy only attempts insertion once it detects a hard impact between the peg and the inner wall of the hole through the release of energy stored in the spring from compression. %Although this is a negative result, it also demonstrates that the policy is truly using feedback from the history of force readings to produce actions instead of executing learned open-loop motions. 

We expect that further reduction of the sim-to-real gap may be achieved through further randomization of key physical parameters such as spring stiffness and damping or surface friction, or through inclusion of randomization parameters in the privileged information. Additionally, rewards can be added to encourage faster task completion to prevent this behavior.
% as done in previous work~\cite{lee2020learning, kumar2021rma}.
Further future work could consider practical insertion scenarios with tighter tolerance parts, or electrical connectors.
% For future work, we will consider more practical insertion, such as more tight tolerance parts or electrical connectors.

%Another limitation of our work is training speed. A major advantage of synthesizing controllers through sim-to-real RL is the potential to leverage massively parallel computations to simulate physics orders of magnitude faster than real life~\cite{rudin2022learning}. Our project did not realize this, as our unoptimized simulation written with MuJoCo python bindings ran on a single CPU thread, resulting in teacher and student training taking approximately 24 and 12 hours, respectively. We hope to explore parallelization options for Robosuite in future work.

%Other possibilities in flexibly defining the privileged data can also be explored, such as including randomized physical parameters or additional task-specific progress information not directly corresponding to physical quantities.

% \addtolength{\textheight}{-1cm}   % This command serves to balance the column lengths
                                  % on the last page of the document manually. It shortens
                                  % the textheight of the last page by a suitable amount.
                                  % This command does not take effect until the next page
                                  % so it should come on the page before the last. Make
                                  % sure that you do not shorten the textheight too much.

%%%%%%%%%%%%%%%%%%%%%%%%%%%%%%%%%%%%%%%%%%%%%%%%%%%%%%%%%%%%%%%%%%%%%%%%%%%%%%%%

%%%%%%%%%%%%%%%%%%%%%%%%%%%%%%%%%%%%%%%%%%%%%%%%%%%%%%%%%%%%%%%%%%%%%%%%%%%%%%%%

%%%%%%%%%%%%%%%%%%%%%%%%%%%%%%%%%%%%%%%%%%%%%%%%%%%%%%%%%%%%%%%%%%%%%%%%%%%%%%%%
% \section*{APPENDIX}

% Appendixes should appear before the acknowledgment.

\section*{ACKNOWLEDGMENTS}
We wish to thank the members of the ETH Zürich Robotic Systems Lab for their valuable discussions.
This study is supported by JST ACT-X, Grant Number JPMJAX22AC.

%%%%%%%%%%%%%%%%%%%%%%%%%%%%%%%%%%%%%%%%%%%%%%%%%%%%%%%%%%%%%%%%%%%%%%%%%%%%%%%%

\bibliographystyle{IEEEtran}
\bibliography{bibliography}

\end{document}